\documentclass[10pt,twocolumn,letterpaper]{article}

 \usepackage{cvpr}              

\definecolor{cvprblue}{rgb}{0.21,0.49,0.74}
\usepackage[pagebackref,breaklinks,colorlinks,allcolors=cvprblue]{hyperref}
\usepackage{graphicx}
\usepackage{amsmath}
\usepackage{amssymb}
\usepackage{booktabs}

\usepackage{algorithm}

\usepackage{epsfig}
\usepackage{enumitem}
\usepackage{bbding} 

\usepackage{times}
\usepackage{caption}
\usepackage{subcaption}
\usepackage{array}
\usepackage{colortbl}
\usepackage{bbm}
\usepackage{multirow}
\usepackage{multicol}
\usepackage{makecell}
\usepackage{xcolor}
\usepackage{xspace}
\usepackage{color}

\usepackage{pifont}

\usepackage{tabulary}
\usepackage{dblfloatfix}
\usepackage{transparent}

\usepackage{listings}
\usepackage{balance}

\definecolor{iGray}{gray}{0.9}
\definecolor{beaublue}{rgb}{0.74, 0.83, 0.9}
\definecolor{Royal_Blue}{rgb}{0.0, 0.1, 0.66}

\usepackage{threeparttable}
\usepackage{longtable}
\usepackage{rotating}
\usepackage{tabularx}
\usepackage{mwe}
\usepackage{bm}
\usepackage{float}
\usepackage{amsfonts}
\usepackage{fixltx2e} 
\usepackage{url}

\usepackage{times}
\usepackage{epsfig}
\usepackage{subcaption}
\usepackage{graphicx}
\usepackage{amsmath}
\usepackage{amssymb}
\usepackage{multirow}
\usepackage{algpseudocode} 

\usepackage{mathtools}

\definecolor{cGreen}{RGB}{78,134,86}
\usepackage[normalem]{ulem}

\usepackage{color, colortbl}
\definecolor{iGray}{gray}{0.9}
\definecolor{beaublue}{rgb}{0.74, 0.83, 0.9}
\definecolor{Royal_Blue}{rgb}{0.0, 0.1, 0.66}
\definecolor{cGreen}{RGB}{100,180,100}

\definecolor{mygray}{gray}{.9}
\newlength\savewidth

\renewcommand{\paragraph}[1]{\vspace{1.25mm}\noindent\textbf{#1}}

\newcolumntype{x}[1]{>{\centering\arraybackslash}p{#1pt}}
\newcolumntype{y}[1]{>{\raggedright\arraybackslash}p{#1pt}}
\newcolumntype{z}[1]{>{\raggedleft\arraybackslash}p{#1pt}}

\newcommand{\app}{\raise.17ex\hbox{$\scriptstyle\sim$}}

\definecolor{deemph}{gray}{0.6}

\definecolor{baselinecolor}{gray}{.9}

\usepackage{url}
\usepackage{graphicx}
\usepackage{multirow}
\usepackage{makecell}
\usepackage{amsmath}
\usepackage{pifont}
\usepackage{hyperref}

\definecolor{myblue}{rgb}{0.0,0.18,0.65}
\hypersetup{ pdftitle={},
pdfauthor={},
pdfsubject={},
pdfkeywords={},
pdfborder=0 0 0,
pdfpagemode=UseNone,
colorlinks=true,
linkcolor=myblue,
citecolor=myblue,
filecolor=myblue,
urlcolor=myblue,    
pdfview=FitH}

\usepackage[capitalize,noabbrev]{cleveref}
\usepackage{appendix}
\usepackage{cleveref}
\crefname{section}{Section}{Sections}
\crefname{theorem}{Theorem}{Theorems}
\crefname{lemma}{Lemma}{Lemmas}
\crefname{equation}{Equation}{Equations}
\crefname{proposition}{Proposition}{Propositions}
\crefname{claim}{Claim}{Claims}
\crefname{appendix}{Appendix}{Appendices}
\crefname{figure}{Figure}{Figs}
\crefname{table}{Table}{Tables}
\crefname{remark}{Remark}{Remarks}
\crefname{definition}{Definition}{Definitions}
\crefname{equation}{Equation}{Equations}
\crefname{corollary}{Corollary}{Corollaries}
\crefalias{section}{appendix}
\definecolor{cGreen}{RGB}{100,180,100}
\definecolor{cRed}{RGB}{220,50,0}

\def\paperID{9383} 
\def\confName{CVPR}
\def\confYear{2025}

\title{Mamba-Adaptor: State Space Model Adaptor for Visual Recognition}

\author{Fei Xie$^1$ \quad Jiahao Nie$^2$\textsuperscript{\thanks {Corresponding author.} } \quad  Yujin Tang$^1$ \quad  Wenkang Zhang$^3$ \quad Hongshen Zhao$^3$ \\
		$^1$~Shanghai Jiao Tong University\\
        $^2$~Hangzhou Dianzi University\\
                $^3$~Southeast University\\
		{\tt\small 	jaffe0319@gmail.com, jhnie@hdu.edu.cn}\\
	}

\begin{document}
\maketitle
\begin{abstract}
Recent State Space Models (SSM), especially Mamba, have demonstrated impressive performance in visual modeling and possess superior model efficiency. However, the application of Mamba to visual tasks suffers inferior
performance due to three main constraints existing in the sequential model: 1) Casual computing is incapable of accessing global context; 2) Long-range forgetting when computing the current hidden states; 3) Weak spatial structural modeling due to the transformed sequential input. To address these issues, we investigate a simple yet powerful vision task Adaptor for Mamba models, which consists of two functional modules: Adaptor-T and Adaptor-S. When solving the hidden states for SSM, we apply a lightweight prediction module Adaptor-T to select a set of learnable locations as memory augmentations to ease long-range forgetting issues. Moreover, we leverage Adapator-S, composed of multi-scale dilated convolutional kernels, to enhance the spatial modeling and introduce the image inductive bias into the feature output. Both modules can enlarge the context modeling in casual computing, as the output is enhanced by the inaccessible features. We explore three usages of Mamba-Adaptor: A general visual backbone for various vision tasks; A booster module to raise the performance of pretrained backbones; A highly efficient fine-tuning module that adapts the base model for transfer learning tasks. Extensive experiments verify the effectiveness of Mamba-Adaptor in three settings. Notably, our Mamba-Adaptor achieves state-of-the-art performance on the ImageNet and COCO benchmarks. 

\end{abstract}

\section{Introduction}
The architecture of Structured State Space Models (SSMs)~\cite{mamba, s4nd, s4, hippo} has gained popularity in language tasks because of their efficiency in modeling long sequences. 
Recently, SSMs, especially the Mamba variant~\cite{mamba}, have been applied to vision tasks after the tremendous success of the vision transformers~\cite{attn, vit, swin, pvt, predformer, towardgsot, sbt, supersbt}. 
While the Mamba-based visual models show potential in various downstream tasks~\cite{dualtfr, videotrack, zhang2024crtrack, samn}, such as dense prediction~\cite{zhu2024vim, vmamba, localmamba, plainmamba}, image synthesis~\cite{dong2024fusionmamba, ma2024umamba}, and generative tasks~\cite{dim, dimba, diffusiontrack}, the further application and adaptation of Mamba models for the vision field remain largely unexplored. 

\begin{figure}[t]
\centering
\includegraphics[scale = 0.40]{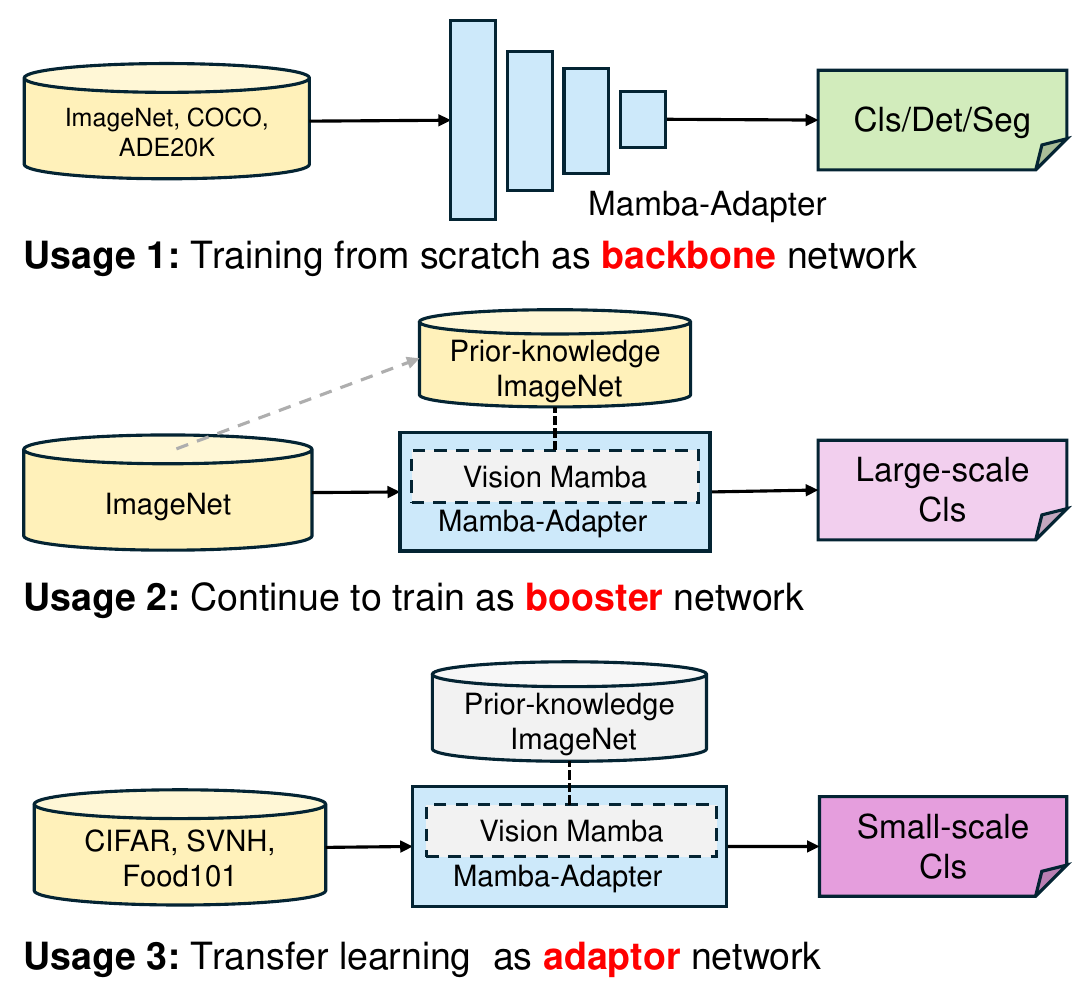}
\caption{Our Mamba-adaptor has multiple usages. It can serve as a general visual backbone network, a booster module for pre-trained backbone, and an adaptor for transfer learning.}
\vspace{-4mm}
\label{fig:tease_usage}
\end{figure}

Despite the superior linear computation complexity with respect to long token length, vision Mamba models have not attained optimal performance in various vision tasks~\cite{ImageNet, coco, ade20k, xiehmt} because of three inherent constraints in the SSM mechanism.
Firstly, the inherent RNN~\cite{lstm} mechanism of SSM recurrently computes the next time step output using the previously stored memory, also dubbed as hidden states. 
Thus, the casual computing pipeline is incapable of accessing future output states, while the 2D visual data is a non-casual style. 
Existing vision Mamba~\cite{zhu2024vim} models partially address this issue with a bi-directional scanning strategy, which reverses the sequential order to modeling global context. 
Secondly, the recurrent computing pipeline in SSM keeps the complexity constant while the influence of the previous hidden state decays as the sequence length increases. 
Thus, the memory decay results in a long-range forgetting phenomenon, which further weakens the global context modeling capability of the SSM. 
Thirdly, the current vision Mamba models~\cite{zhu2024vim,vmamba} heavily rely on transforming 2D visual data into 1D sequential format. which destroys the spatial dependencies in structural images.
Though dedicated scanning strategies, such as four-way~\cite{vmamba} and local windowed scanning~\cite{plainmamba, localmamba}, can partially preserve the spatial localities, the vision Mamba models still suffer from learning 2D dependencies from the sequence modeling.

\begin{figure}[t]
\centering
\includegraphics[scale = 0.42]{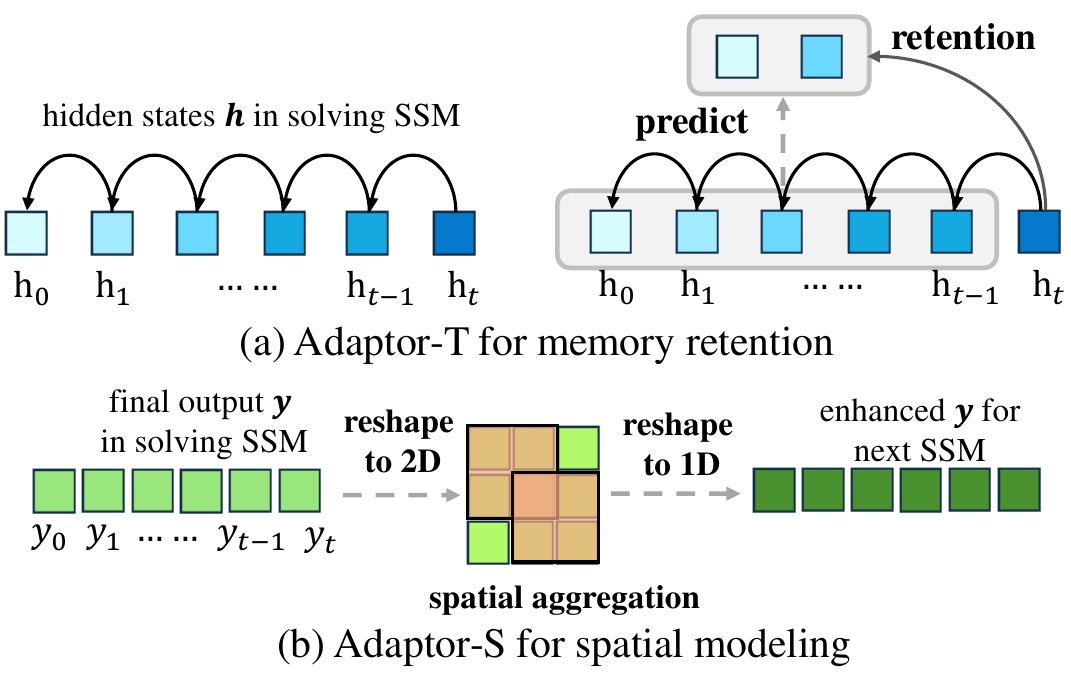}
\caption{Mamba-Adaptor consists of two modules for solving SSM~\cite{mamba, s5}: Adaptor-T conducts memory retention for the hidden states; Adaptor-S conducts spatial aggregation for final output.}
\vspace{-4mm}
\label{fig:tease-adaptor}
\end{figure}

After a detailed analysis, we classify the three mentioned constraints into two categories, i.e., temporal decay and spatial locality. 
Therefore, we introduce a simple yet powerful vision task adaptor for Mamba models (shown in Figure~\ref{fig:tease-adaptor}), which consists of two functional modules: Adaptor-T and Adapator-S.
Adaptor-T conducts memory retention to handle the long-range forgetting issue for each hidden state. 
It is noteworthy that handcrafted memory selection falls short in finding the most forgetting states, limiting the effects of memory retention. 
Thus, we apply a lightweight learnable memory selection scheme that dynamically selects a group of the most forgetting states. 
For spatial modeling enhancement, we directly apply multi-scale depthwise convolution filters to functions at the 2D feature output, which effectively introduces the image inductive bias. 
Moreover, it is critical to implement our adaptor into the highly optimized Mamba solver. 
To achieve this, we decouple the solver with identity matrix and highly efficient matrix multiplication, thus adaptor-S and Adpater-T can be inserted into Mamba solvers seamlessly. 
We also devise two insertion forms for Mamba-Adaptor, i.e., parallel form for transfer learning and sequential form for training from scratch.

As illustrated in Figure~\ref{fig:tease_usage}, our Mamba-Adaptor has multiple usages, including serving as a general vision backbone, a booster module to raise the performance further, and an efficient Adaptor for transfer learning. 
Mamba-Adaptor achieves superior performance compared to recent Mamba-based, CNN, and transformer-based methods in classification and dense prediction tasks.
Mamba-Adaptor also shows competitive results on various transfer learning tasks. 
Overall, the contribution is as follows:

\begin{itemize}
	\item We introduce a plug-and-play Mamba-Adaptor, which can enhance the performance of Mambal models in various vision tasks with acceptable computational overhead.

	\item Our Mamba-adaptor consists of two functional modules, Adaptor-T and Adaptor-S, which effectively address the temporal decay and spatial locality issues. 
  
  	\item Mamba-Adaptor not only can serve as a general vision backbone for image classification and prediction tasks but also take effects in transfer learning task settings.  
   
	\item Extensive experimental results show our Mamba-Adaptor's superior performance and considerable improvements over Mamba baselines across various vision downstream tasks.
\end{itemize}





\section{Related Work}


\paragraph{State Space Model.}
State Space Models (SSMs)~\cite{s4,s5,s4nd, mamba}, originating from linear control systems~\cite{ssm}, offer an efficient framework with linear computational complexity that makes them well-suited for long-sequence processing. 
Language tasks~\cite{mehta2023long, glue} are first applied with the SSM-based methods for long-range context modeling.
Later, a bunch of improvements, including structural optimizations~\cite{s4}, associative scannings~\cite{s5}, and hardware speed-up techniques~\cite{hippo}, are proposed to evolve the SSM with modern deep learning architecture~\cite{alexnet, googlenet,inceptionnext, ResNet, resnext, Pytorch,fast_rcnn, mask-rcnn}. 
The recently emerged Mamba~\cite{mamba} introduced input-specific parameterization and parallel scanning (S6), further establishing SSMs as a strong competitor to Transformers.
These innovations have led to the widespread application of SSMs in vision tasks, especially as an alternative for transformers~
\cite{swin, pvt, pvtv2, crossformer, deit, davit} with superior linear computation complexity.
Pioneer vision models such as ViM~\cite{zhu2024vim} and VMamba~\cite{vmamba} apply the Mamba scheme in visual representation by transforming the 2D images into a sequential format using bidirectional- and cross-scanning strategies.
The following vision methods improve the visual capability of Mamba scanning through a more dedicated scanning pattern~\cite{localmamba, plainmamba, pei2024efficientvmamba, quadmamba} to preserve spatial locality and hybrid transformer-Mamba architecture~\cite{mambavision, ma2024umamba} as powerful complements. 
Despite these advances, visual SSM models still face challenges in overcoming the long-range modeling issues and sequential transformation limitations for processing 2D image data, which impedes their performance for vision tasks that require vision-specific priors and capabilities. 
Our adaptor aims to overcome the shortcomings of long-range forgetting issues and sequence transformation in vision Mamba-based methods.

\paragraph{Visual adaptor.}
The concept of the adaptor originates from the language field~\cite{lora}, which functions as an additional module for fine-tuning large language models for specific purposes~\cite{houlsby2019parameter, pals2019}.
For the vision field, adaptors generally serve as trainable parts to transfer the large-scale knowledge prior to vision foundation models~\cite{internimage, BERT, CLIP, swin-v2}, such as CLIP~\cite{CLIP}, into the various downstream tasks~\cite{cheng2023meta, zhang2021tip, sung2022vl, gao2024clip}.
For transfer learning tasks~\cite{transferlearning, prompt, lora}, Adaptformer~\cite{adaptformer} uses an additional branch for MLP~\cite{attention} to finetune the downstream tasks for the pre-trained vision transformer efficiently. 
VPT~\cite{prompt} adds extra visual tokens as prompt input to fine-tune the base model efficiently. 
The transfer learning~\cite{transferlearning} for Mamba models still remains largely unexplored. 
We develop Mamba-specific fine-tuning modules to adapt Mamba to downstream tasks for transfer learning. 
Another usage of adaptors in vision is as a complementary module to enhance the inability of the base model, such as enhancing the hierarchical feature representation in plain vision transformers~\cite{vit}. 
Pioneer works like ViTDet~\cite{li2022vitdet} implements feature pyramid modules to adapt the plain features extracted from the vision transformer for dense prediction tasks. 
Another representative work in visual adaptor is ViT-adaptor~\cite{vit-adapter}, which designs a convolution-based network as a complement branch to the attention-based network, which introduces the necessary inductive bias for image tasks.  
The Mamba-based architecture~\cite{s4, s5, mamba} is originally designed for language modeling, which neglects the visual priors. 
We introduce the visual priors and overcome the Mamba constraints, which can serve as a booster module. 
Our work explores two usages of the Mamba-adaptor, which serves as a dedicated adaptor design for Mamba models in transfer learning, and a booster module to raise the performance further.

\section{Method}
\label{sec:method}

In this section, we first revisit the preliminaries of the SSM in Section~\ref{sec:pre}. 
Then, we introduce the formulation of the proposed Mamba-Adaptor (Section~\ref{sec:adaptor}) and the two vital submodules: Apdator-T (Section~\ref{sec:adaptor-T}) and Adaptor-S (Section~\ref{sec:adaptor-S}). 
We also depict the practical implementations of the Mamba-Adaptor and how to integrate the adaptor into the Mamba structure in Section~\ref{sec:implem}.

\subsection{Preliminaries}
\label{sec:pre}

\begin{figure*}[t]
\centering
\includegraphics[scale = 0.42]{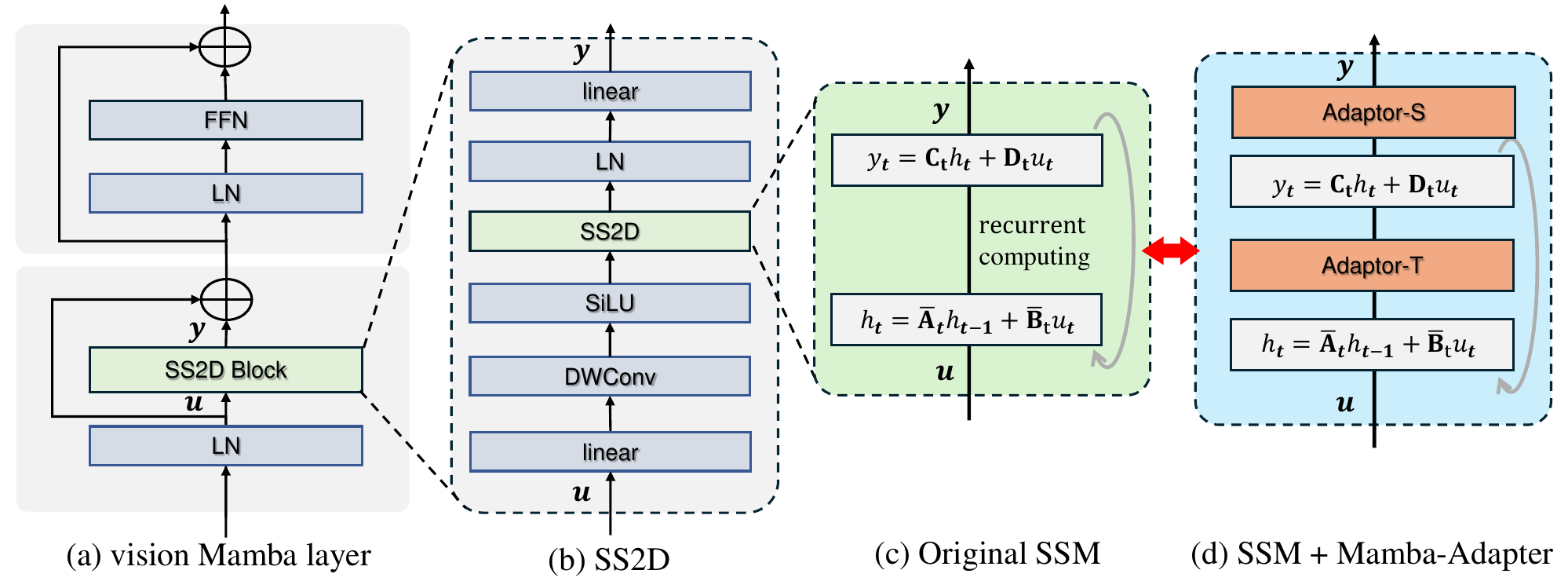}
\vspace{-2mm}
\caption{Details of the Mamba-Adapter. (a) illustrates a typical architecture of vision Mamba layer~\cite{vmamba}, including SS2D block, FFN~\cite{attention} and two residual connections; (b) details of SS2D layer, which transforms the 2D image into sequences and process by SSM equation; (c) recursive solving procedure of SSM with two main steps; (d) Mamba-Adaptor with Adaptor-T and Adaptor-S in solving SSM process. FFN and LN denote feed-forward network~\cite{attn} and layer normalization~\cite{LN}.}
\vspace{-4mm}
\label{fig:Adaptor}
\end{figure*}

\paragraph{State Space Model.}
State-space models (SSMs) are often utilized to analyze sequential data and model the in-out transform for continuous linear time-invariant (LTI) systems.
In SSM, an input sequence $u(t) \in \mathbb{R}$ is to compute the response $y(t) \in \mathbb{R}$ by the hidden state $\mathbf{h}(t) \in \mathbb{R}^{N}$ as an intermediate value.
The process of solving SSM can be expressed as a combination of linear ordinary differential equations (ODEs):
\begin{equation}\label{eq:ssm}
\begin{split}
  \mathbf{h'}(t) &= \mathbf{A} \mathbf{h}(t) + \mathbf{B} u(t),  \\
  y(t)  &= \mathbf{C} \mathbf{h}(t) + D u(t),
\end{split}
\end{equation}
where $\mathbf{A} \in \mathbb{R}^{N\times N}$, $\mathbf{B}\in \mathbb{R}^{N\times 1}$, $\mathbf{C}\in \mathbb{R}^{1\times N}$, and $D \in \mathbb{R}^{1}$ are the transforming matrices for the linear state transition and output equations.

\paragraph{Discretization of SSM.}\label{sec:discretization_of_ssm}
To effectively solve the SSM~\cite{mamba} in the LTI system within a deep learning framework, it is important to convert the continuous-time equations into a discrete format. 
A frequently used method in LTI analysis for achieving this is called Zero-Order Hold (ZOH) discretization~\cite{mamba-nd}.
A continuous-time system can be converted into a discrete-time system by breaking the input into small time intervals, which can be considered as constant during those intervals.
ZOH effectively converts continuous-time parameters ($\mathbf{A}$, $\mathbf{B}$, $\mathbf{C}$, $\mathbf{D}$) in SSM into their discrete equivalents over the fixed sampling time interval $\Delta \in \mathbb{R}^D$:
\begin{equation}
\begin{aligned}
\bar{\mathbf{A}}&=e^{\Delta \mathbf{A}}, \quad \bar{\mathbf{B}}=\left(\Delta\mathbf{A}\right)^{-1}\left(e^{\Delta \mathbf{A}}-I\right)  \Delta\mathbf{B},\\
\bar{\mathbf{C}}&= \mathbf{C}, \quad ~~~~\mathbf{D}=\mathbf{D}.\\
\end{aligned}
\end{equation}

\paragraph{Selective Scan Mechanism.}\label{sec:selective_scan_mechanism}
To address the limitations of linear time-invariant (LTI) state-space models (SSMs) in capturing contextual information, Mamba~\cite{mamba} introduced a parameterization method featuring an input-dependent selection mechanism called S6~\cite{s4}. This allows the model to select relevant features based on specific inputs, enhancing its response to various input contexts.
In the context of selective SSMs, the time-varying weighting parameters pose a challenge for efficiently calculating hidden states since convolutions are static and are not designed to handle dynamic weights. 
Mamba~\cite{gu2023mamba} applys an input-dependent scheme to generate the paramters $\{\mathbf{B}_{k}\}_{k=1}^{L} $, $\{\mathbf{C}_{k}\}_{k=1}^{L} $ and $\{\mathbf{\Delta}_{k}\}_{k=1}^{L} $ from the input sequence $\{x_{k}\}_{k=1}^{L} $.
With the discretization of paramters, the sequential states $\{h_{k}\}_{k=1}^{L}$ can be computed in a recursive manner:
\begin{equation}
\label{equa:ssm_delta1}
\begin{aligned}
h_{t} &=\bar{\mathbf{A}}_{t} h_{t-1}+\bar{\mathbf{B}}_{t} u_{t},\\
\end{aligned}
\end{equation}
The hidden state $h_t$ in Eq.~\ref{eq:ssm} can be computed recursively using the previous hidden state $h_{t-1}$ and current feature input $u_t$. Then, we can obtain the final output $y_t$ from the hidden state:
\begin{equation}
\label{equa:ssm_delta2}
\begin{aligned}
y_{t} &=\mathbf{C}_{t} h_{t}+\mathbf{D} u_{t}.\\
\end{aligned}
\end{equation}
The response $y_b$ can still be efficiently computed using associative scan algorithms~\cite{s5} with linear complexity. Hardware optimization tricks are also applied to speed up solving SSM~\cite{mamba, s4}.

\paragraph{Sequence Transform for Vision Mamba.} 
Vision Mamba methods~\cite{zhu2024vim, vmamba, localmamba} has to convert the 2D image data  $x\in \mathbb{R}^{H\times W}$ into the sequential format $x_s\in \mathbb{R}^{L=H\times W}$ in advance. 
To gain global context, bidirectional scanning is the basic strategy, while multiple scanning routes are encouraged to obtain more through context modeling ability.
For multiple scanning sequences, SSM processes each of them independently and merges the outputs in the final. 
Here we name the sequence transform and SSM solving procedure in vision Mamba as SS2D scheme.

\subsection{Formulation of Mamba-Adaptor}
\label{sec:adaptor}
Our Mamba-Adaptor directly functions at the solving Equation~\ref{equa:ssm_delta1} and Equation~\ref{equa:ssm_delta2} of SSM, which aims to enhance the modeling capability inherently.
We explore the potential limitations of the two main steps, i.e., hidden state computation in Equation~\ref{equa:ssm_delta1}  and output computation in Equation~\ref{equa:ssm_delta2} of solving SSM.

\paragraph{Hidden State.}
Each hidden state $h_i$ for index $i$ embedding in Equation~\ref{equa:ssm_delta1} is a summation of the previous hidden state $h_{t-1}$ and the transformed value of current input $u_t$.
The hidden states are computed recursively, beginning with an empty state $h_{0}$. 
Thus, the far-away hidden state is inaccessible to the current hidden state $h_t$ when solving Equation~\ref{equa:ssm_delta1}, resulting in a long-range forgetting issue.  
Being aware of this issue, a weighted aggregation of hidden states from the non-neighboring area can effectively mitigate the forgetting:
\begin{equation}
\label{equa:forget}
\begin{aligned}
h_{t} &=  h_{i} + \sum_{\forall j \in \Lambda}( w_{j} h_{j}),\\
\end{aligned}
\end{equation}
where $\Lambda$ is an index set of selected accessible hidden states and $w$ is the corresponding weight parameters for summation. 
Equation~\ref{equa:forget} formulates the core principle of Adaptor-T, which handles the temporal decay issue in the hidden state of Mamba.

\paragraph{Final Output.}
In  Equation~\ref{equa:ssm_delta2}, the final transformed output $y_t$ is obtained by a summation of transformed hidden state $h_t$ and corresponding input embedding $u_t$. 
Then, the final sequential output $ \{y_0, y_1, ..., y_L\} $ is reshaped to 2D format to obtain the final output $\mathbf{y}\in \mathbb{R}^{H\times W}$ for the next layer. 
Though the $h_t$ contains the long-range dependencies through recursive hidden state computation, the spatial locality in 2D image data is largely overlooked in the final output.
To overcome this issue, we function at the reshaped 2D format output $\mathbf{y}$ to obtain the spatial dependencies through a Convolutional-style aggregation, which can introduce the image inductive bias effectively: 
\begin{equation}
\label{equa:conv}
\begin{aligned}
 \mathbf{y} &:=  \text{Conv}(\mathbf{y}),\\
\end{aligned}
\end{equation}
where $\text{Conv}$ operator aggregates the weighted output $\mathbf{y}$ in a fixed scanning local window. 
For each output $y_{(i,j)}$ in 2D spatial coordinate $(i,j)$, the spatial locality is enhanced as follows:
\begin{equation}
\label{equa:conv_each}
\begin{aligned}
y_{(i, j)} &=  \sum_{\forall (i, j) \in \Omega} w_{(i,j)} y_{(i, j)},\\
\end{aligned}
\end{equation}
where $\Omega$ is generally $K\times K$ local neighbouring embeddings and $w$ is the corresponding weight parameters.
The local adjacency can be enlarged by increasing the kernel size $K$. 
Equation~\ref{equa:conv} and Equation~\ref{equa:conv_each} formulate the core scheme of our Adaptor-S, which functions at the spatial locality of the output in  Mamba.

\begin{algorithm}[t]
\caption{Pseudo code of Mamba-Adaptor}
\label{algo:Mamba-Adaptor}
\definecolor{codeblue}{HTML}{2E8B57} 
\definecolor{codekw}{HTML}{DC143C} 
\lstset{
  backgroundcolor=\color{white},
  columns=fullflexible,
  breaklines=true,
  captionpos=b,
  commentstyle=\fontsize{7.2pt}{7.2pt}\color{codeblue},
  keywordstyle=\fontsize{7.2pt}{7.2pt}\color{codekw},
  escapechar={|}, 
}
\lstset{language=Python}
\begin{lstlisting}[xleftmargin=-1em]
 #input: u; params: delta, A, B, C, D; output: y
 def Mamba-Adaptor(images, maps):
     # Generating identity Matrix for C
     C_identiy = torch.ones_like(C)
     # Generating zero Matrix for C
     D_zeros = torch.zeros_like(D) 
     # Calculate hidden state using Mamba solver
     hidden_state = SelectiveScanCuda(u, A, B, C_identiy, D_zeros, delta)
     # Adaptor-T
     hidden_state = Adaptor_T(hidden_state)
     # Calculate output using matrix multiplication
     y = C*hidden_state + D*u
     # Adaptor-S
     y = Adaptor_S(y)
     return y
\end{lstlisting}
\end{algorithm}

\subsection{Adaptor-T}
\label{sec:adaptor-T}
The aim of Adaptor-S is to conduct memory retention for each hidden state $h_i$. 
Since Equation~\ref{equa:ssm_delta1} is casually computed, each hidden state only has access to the previous states, whose sequential order is smaller than the current one. 
Brutally aggregating all the previous states may not work well since the forgetting degrees of previous hidden states vary. 
A fixed handcrafted pattern to select neighboring hidden states, which acts like convolutional filters, is also insufficient, as the hidden state that decays may be largely overlooked.

\paragraph{Learnable Memory Selection.}
To handle this, we propose a learnable memory selection scheme, which uses an extremely lightweight prediction layer to select a set of most forgetting hidden states. 
We adopt an extremely lightweight linear layer $\phi_p(.)$ to predict $K$ coordinates $p_k$ to select the forgetting states for each hidden state $h_i$. The coefficient of each forgetting state $c_k$ is also predicted by a linear layer $\phi_c(.)$ with a SoftMax function~\cite{attention}:
\begin{equation}
\begin{aligned}
 \{p_0, p_1, ..., p_k\} &= \phi_p(h_i),\\
  \{c_0, c_1, ..., c_k\} &= \text{SoftMax}(\phi_c(h_i)),\\
\label{eqn:dcn1} 
\end{aligned}
\end{equation}
Therefore, the memory states most likely to be forgotten are dynamically determined, eliminating the need for manual selection.

\paragraph{Multi-Sequence Temporal Retention.}
Similar to the Multi-head attention scheme~\cite{vit} and multi-group in convolution filters, the total $S$ sequences are separately predicted and aggregated:
\begin{equation}
\begin{aligned}
    h_i^s &:= \sum^{K}_{k=1} \mathbf{c}_{sk}\mathbf{h}_s(p_k), s={1,2,\dots,S}\\
        & w.r.t~~~  \sum^{K}_{k=1} \mathbf{c}_{sk} = 1.\\
\label{eqn:dcn2} 
\end{aligned}
\end{equation}
For the bidirectional scanning~\cite{zhu2024vim}, the total number of the sequence $S$ is 2, which needs to predict $2K$ coordinates. 
Thus, each hidden state in SSM is strengthened by the predicted most forgetting states, similar to the memory retentive scheme in language processing.

\subsection{Adaptor-S}
\label{sec:adaptor-S}
Guided by the principle of Adaptor-S, 
we adopt depthwise convolution filters to aggregate the output features $\mathbf{y}$ from the 2-dimensional spatial aspect.
The depthwise convolution is lightweight, bringing acceptable computation overhead. 
It aggregates the features channel-wise, thereby keeping the shape of output features unchanged.

\begin{figure}[t]
\centering
\includegraphics[scale = 0.45]{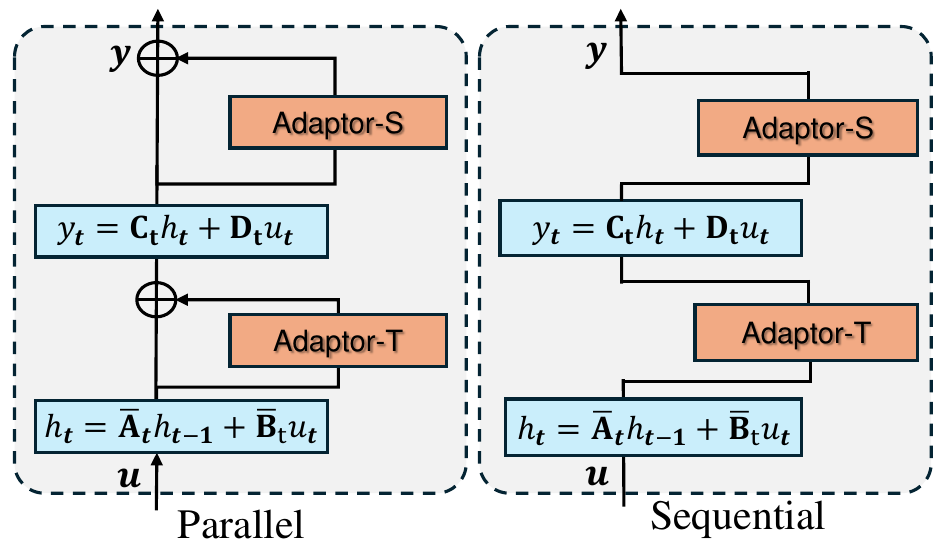}
\caption{Illustration of the parallel
and sequential insertion form of Mamba-Adaptor.}
\vspace{-4mm}
\label{fig:seq}
\end{figure}

\paragraph{Multi-Scale Spatial Aggregation.}
Specifically, to enlarge the spatial dependencies, we use multiple depth-wise convolutional filters with multi-scale dilation factors. 
The spatial locality strengthened for final output feature embedding $y_(i, j)$ in spatial coordinate $(i, j)$ can be expressed as follows:
\begin{equation}
\label{equa:depthwise}
\begin{aligned}
y_{(i, j)} &=  \sum_{\forall d } \sum_{\forall (i, j) \in \Omega_{d}} w^{d}_{(i,j)}  y_{(i, j)},\\
\end{aligned}
\end{equation}
where $\Omega_{d}$ is the index selected from the neighboring area of the depthwise convolutional filter $d$.  
This spatial aggregation process in Adaptor-S brings more structural inductive bias to the sequential feature representations, greatly enhancing the spatial locality in Mamba.

\subsection{Implementation}
\label{sec:implem}
We describe how to implement the Mamba-Adaptor using highly optimized hardware modules and introduce two integrated methods for inserting adaptors tailored to different task settings.

\paragraph{Practical Implementation.}
Integrating the Mamba-Adaptor into the highly optimized SSM implementation is essential for effectiveness.
We separate the solving procedure into two parts: an original, optimized Mamba operator and highly efficient matrix multiplications.
Our Adaptor-T and Adaptor-S can be easily integrated into the Mamble layer.
Concretely, we manually generate an identity matrix and zero matric for $\mathbf{C}$ and $\mathbf{D}$, respectively, to replace the original parameter. 
Therefore, the highly optimized Mamba solver~\cite{mamba} can compute the hidden states instead of the final output, which is where Adaptor-T can be inserted. 
Then, we use matrix multiplications to calculate the final output with Adaptor-S to enhance the spatial modeling.
The overall procedure can be described in the pseudo-code in Alg.~\ref{algo:Mamba-Adaptor}.

\paragraph{Layer Insertion}
\label{sec:Integration}
As illustrated in Figure~\ref{fig:seq}, we introduce two forms to integrate our adaptor with vision Mamba models, i.e., sequential and parallel insertion.
For image classification and dense prediction tasks that are trained from scratch, we apply sequential form. 
For transfer learning which finetunes a pretrained model, we adopt the parallel form.
The parallel design preserves the original pretrained features through an independent branch, which is better suited for fine-tuning in transfer learning tasks.

\paragraph{Weight-Sharing Coefficient.}
For transfer learning and booster settings, which need a zero weight initialization and less fine-tuned parameters to achieve high efficiency, 
we adopt a  weight-sharing coefficient strategy for Adaptor-S. 
Concretely, we initialize fixed-shape parameters for coefficients $\{c_0, c_1, ..., c_k\}$ for each hidden state, thus reducing the model cost of linear layer $\phi_c$.


\section{Experiment}
We evaluate the effectiveness of the Mamba-Adaptor by three major experimental settings.
The experiment in Section~\ref{sec:backbone_exp} assesses the performance of the visual backbone network for both image classification and dense prediction tasks. 
The experiment in Section~\ref{sec:booster_exp} involves using the Mamba-Adaptor as a booster module to enhance the performance of pre-trained networks. 
The experiment in Section~\ref{sec:adaptor_exp} 
 focuses on transfer learning, where the Mamba-Adaptor acts as an efficient fine-tuning module for downstream recognition tasks. 
%
%
In the final, we depict the ablation studies in Section~\ref{sec:abla_exp}.

\subsection{Visual Backbone Network}
\label{sec:backbone_exp}

\paragraph{Model Architecture.}
To serve as a visual backbone network, Mamba-Adaptor follows a widely-adopted hierarchical backbone architecture~\cite{ResNet, xie2017resnext, huang2017densenet, swin, pvt, vmamba}. 
It primarily consists of a patch embedding layer, a vision mamba layer featuring the proposed Mamba-Adaptor to form four model stages, and a head network for classification and dense prediction.
Each model stage includes a downsampling layer that reduces the size of feature maps by half while increasing their channel dimension twofold.
In this experiment, we introduce two variants of the Mamba-Adaptor, which are as follows:

Mamba-Adaptor-b1 with channel dimension nof $48$ and layer number $\{2, 2, 6, 2\}$ in four model stage; Mamba-Adaptor-b2 with channel dimension of $96$ and layer number $\{2, 2, 8, 2\}$ in four model stage.

\begin{table}[t]
    \centering
    \setlength{\tabcolsep}{3.8mm}
    \subfloat{
    \scalebox{0.80}{
    \begin{tabular}{c | c c |c}
        \toprule[1pt]
          Method & \makecell{Parmas\\(M)} & \makecell{FLOPs\\(G)} & \makecell{Top1-acc\\(\%)}\\
        \midrule[0.5pt]
        RegNetY-1.6G & 11  & 1.6 & 78.0  \\
        EfficientNet-B3 & 12  & 1.8 & 81.6  \\
        PVTv2-b1~\cite{pvtv2} & 13 & 2.1 & 78.7 \\
        BiFormer-T~\cite{biformer} & 13 & 2.2 & 81.4 \\
        Vim-T &7  & 1.5 & 76.1 \\
        LocalViM-T & 8 & 1.5 & 76.2 \\
         \rowcolor{gray!20} Mamba-Adaptor-b1 & 7.8 & 1.4 & 78.4 \\
        \midrule[0.5pt] 
        CoAtNet-T~\cite{convnext} & 29 & 4.5 & 82.1 \\
        UniRepLKNet-T~\cite{convnext} & 29 & 4.5 & 82.1 \\
        ConvNeXt-T~\cite{convnext} & 29 & 4.5 & 82.1 \\
        MambaoutOut-T & 27 & 4.5 & 82.7 \\
        InternImage-T~\cite{internimage} & 30 & 5.0 & 83.5 \\
        DeiT-S~\cite{deit} & 22 & 4.6 & 79.9 \\
        Swin-T~\cite{swin} & 29 & 4.5 & 81.3 \\
        Vim-S & 26 & 5.1 &80.3  \\
        VMamba-T & 22 & 5.6 & 82.6 \\
        LocalVMamba & 26 & 5.7 & 82.7 \\
         \rowcolor{gray!20} Mamba-Adaptor-b2 & 32 & 5.4 & 83.0 \\
        \bottomrule[1pt]
    \end{tabular}}}
        \vspace{-2mm}
    \caption{Comparison of state-of-the-art methods for ImageNet-1K~\cite{ImageNet} classification.}
    \vspace{-6mm}
    \label{tab:ImageNet}
\end{table}

\begin{table*}[t]
    \centering
    \setlength{\tabcolsep}{2.3mm}
    \scalebox{0.80}{
    \begin{tabular}{c|c c|c c c c c c | c  c c c c c}
        \toprule[1pt]
         \multirow{2}{*}{Backbone} & \multirow{2}{*}{\makecell{Params\\(M)}} & \multirow{2}{*}{\makecell{FLOPs\\(G)}} & \multicolumn{6}{c}{Mask R-CNN $1\times$+MS} & \multicolumn{6}{c}{Mask R-CNN $3\times$+MS}\\
          & & & $AP^b$ & $AP^b_{50}$ & $AP^b_{75}$ & $AP^m$ & $AP^m_{50}$ & $AP^m_{75}$ & $AP^b$ & $AP^b_{50}$ & $AP^b_{75}$ & $AP^m$ & $AP^m_{50}$ & $AP^m_{75}$\\
          \midrule[0.5pt]
          ResNet18  &21.3 & 189 &31.8 &49.6 &33.6 &16.3 &34.3 & 43.2 &-  &-  &-  &- &- &-\\
         PVT-T & 33 & 208 &36.7 &59.2 &39.3 &35.1 &56.7 &37.3 & 39.8 &62.2& 43.0 &37.4 &59.3 &39.9\\
         EffVMamba-S &31 &197 &39.3 &61.8 &42.8 &36.7 &58.9 &39.2 &41.6 &63.9 &45.6 &38.2 &60.8 &40.7\\
          \rowcolor{gray!20}Mamba-Adaptor-b1 & 32 & 218 &43.2 & 65.5 & 47.7 & 39.5 & 60.1 & 42.7 &45.1  &67.2  &49.4  &-41.2&61.9 &43.8 \\
          \midrule[0.5pt]
          ResNet-50 &44 &260 &38.2 &58.8 &41.4 &34.7 &55.7 &37.2 &-  &-  &-  &- &- &-\\
          Swin-T  &48 &267 &42.7 &65.2 &46.8 &39.3 &62.2 &42.2& 46.0 &68.1 &50.3 &41.6 &65.1 &44.9\\
          VMamba-T &50 &271 &46.5 &68.5 &52.0 &42.1 &65.5 &45.3 &48.8 &70.4 &53.5 &43.7 &67.4 &47.0
\\
          LocalVMamba-T &45 &291 &46.7 &68.7 &50.8 &42.2 &65.7 &45.5 &48.7 &70.1 &53.0 &43.4 &67.0 &46.4 \\
          \rowcolor{gray!20}Mamba-Adaptor-b2 & 42 & 259 &47.3 & 69.8 & 52.3 & 43.4 & 66.9 & 46.4  &49.1  &71.5  &54.1  &44.8 &67.3 &48.3\\
          \bottomrule
    \end{tabular}}
            \vspace{-2mm}
    \caption{Comparison to the state-of-the-art backbone networks using Mask R-CNN with "$1\times$" and "$3\times$" training schedules.}
\vspace{-4mm}
    \label{tab:COCO1x}
\end{table*}

\paragraph{Image Classification on ImageNet-1K.}
The ImageNet-1K~\cite{ImageNet} dataset comprises 1.28 million images designated for training and 50K images allocated for validation. 
By using the same training settings~\cite{swin, vmamba}, we train two variants of Mamba-Adaptor from scratch using the training set and evaluate their performance by reporting the Top-1 accuracy on the validation set.
We assessed VMamba's performance in image classification on the ImageNet-1K dataset, and the results comparing it with benchmark methods are presented in Table~\ref{tab:ImageNet}.
With similar computation FLOPs, our Mamba-Adaptor-b1 achieves a top-1 accuracy of 78.4, surpassing LocalViM-T by 2.2 \% and Vim-T by 2.3 \%. 
Moreover, Mamba-Adaptor-b2  also retains its performance advantage at larger scales.
For example,  Mamba-Adaptor-b2 achieves a top-1 accuracy of 82.9, exceeding Swin-T by 2.6 \% and VMamba-T by 0.2\%.

\paragraph{Object Detection on COCO.}
COCO~\cite{coco} is an object detection and instance segmentation dataset that comprises 118K training images and 5K validation images. 
We utilize Mamba-Adaptor as the backbone in MaskRCNN detection framework~\cite{mask-rcnn} to assess the effectiveness of our method.
We follow similar training strategies in SwinT~\cite{swin}, which use the pretrained weights on the ImageNet-1K dataset. 
The results on MSCOCO are shown in Table~\ref{tab:COCO1x}. Our Mamba-Adaptor demonstrates superiority in both box and mask Average Precision ($AP^b$ and $AP^m$), in both "$1\times$" and "$3\times$" training schedules.
With the "$1\times$" fine-tuning schedule and similar computational FLOPs, 
Mamba-Adaptor-b1 achieves best object detection mAPs of 43.2\%, surpassing EffVMamba-S~\cite{pei2024efficientvmamba} by 3.9 \% $AP^b$, and PVT-T by 6.5 \% $AP^m$, respectively. 
Under the fair comparisons, the instance segmentation mAPs achieved by Mamba-Adaptor-b2 outperform the baseline VMamba-T by 1.8\% in $AP^b$ and 2.3 in $AP^m$. 
Mamba-Adaptor-b2 also shows advantages to the improved vision Mamba model, LocalVMamba-T in both object detection and instance segmentation tasks. 
The Mamba-Adaptor demonstrates its potential to deliver impressive performance in downstream tasks related to dense prediction.


\begin{table}[t]
    \centering
    \setlength{\tabcolsep}{2.5mm}
    \subfloat{
    \scalebox{0.78}{
    \begin{tabular}{c | c c |c}
        \toprule[1pt]
          Method & \makecell{Parmas\\(M)} & \makecell{FLOPs\\(G)} & \makecell{Top1-acc\\(\%)}\\
        \midrule[0.5pt]
        VMamba-T & 30 & 4.9 & 82.6 \\
         \rowcolor{gray!20} Adaptor-VMamba-T & 31 ($3.2\% \uparrow $) & 5.2  ($6.1\% \uparrow $) &  \textbf{82.7}\\
        \midrule[0.5pt] 
        VMamba-S &  50& 8.7  & 83.6  \\
         \rowcolor{gray!20} Adaptor-VMamba-S & 53 ($6.1\% \uparrow $)& 9.3 ($8.0\% \uparrow $)& \textbf{83.7}\\
        \midrule[0.5pt] 
        VMamba-B  &  89&  15.4& 83.9 \\
         \rowcolor{gray!20} Adaptor-VMamba-B  & 94 ($5.6\% \uparrow $) & 16.5 ($7.1\% \uparrow $) &  \textbf{84.1}\\
        \bottomrule[1pt]
    \end{tabular}}}
            \vspace{-2mm}
    \caption{Improvements over ImageNet-1K classification. Adaptor-Baseline denotes the baseline model, which is equipped with our Mamba-Adaptor as a booster module. }
    \vspace{-4mm}
    \label{tab:booster_cls}
\end{table}

\subsection{Booster Network for Image Recognition}
\label{sec:booster_exp}

\paragraph{Booster, Baseline, and Benchmark.}
We choose VMamba-T/S/B pretrained in ImageNet~\cite{ImageNet} classification as the baseline model and equip it with the proposed Mamba-Adaptor with acceptable computation overhead. 
We are continuing to fine-tune the VMamba equipped with the Mamba-Adaptor on the ImageNet training set for an additional 10 epochs.

\paragraph{Resutls on Booster.}
The results in Table~\ref{tab:booster_cls} show that our Mamba-Adaptor can raise the performance of the pretrained VMamba~\cite{vmamba} baseline further. 
Concretely, with less than 3.2\% extra model parameters and 6.1 \% FLOPS, Mamba-Adaptor can raise the VMamba-T by 0.1\% top1 accuracy. 
With the VMamba-B with a large model size, the boosting performance is increased to 0.2 \% in top1 accuracy. 
It is worth noting that the booster training only requires 10 additional epochs, which indicates that our Mamba-Adaptor as a booster module is highly efficient.


\begin{table*}[t]
    \centering
    \setlength{\tabcolsep}{4.5mm}
    \subfloat{
    \scalebox{0.78}{
    \begin{tabular}{c | c  c | c | c c c c  c c }
        \toprule[1pt]
         Adaptor Method& Base Model & Pretrain  & \makecell{Parmas\\(M)} & \makecell{CIFAR-100\\(Acc\%)} & \makecell{SVHN\\(Acc\%)} & \makecell{Food101\\(Acc\%)}\\
        \midrule[0.5pt]
        Full-Tuning  & ViT-B~\cite{vit} &MAE~\cite{mae} & 86.04 (100\%) & 85.90& 97.67 &90.09 \\
        Linear & ViT-B~\cite{vit} &MAE~\cite{mae} &  0.07 (0.08\%) &69.83 (-16.07) &66.91 (-30.76) & 69.74 (-20.35)  \\
          VPT~\cite{prompt} & ViT-B~\cite{vit} &MAE~\cite{mae} &  0.08 (0.09\%) &82.44 (-3.46) &94.02 (-3.65) & 82.98 (-7.11)  \\
        \midrule[0.5pt] 
        Full-Tuning &VMamba-T~\cite{vmamba} &Cls.~\cite{vmamba}  & 30.25 (100\%) & 87.48   &97.82 & 90.22\\
         Linear & VMamba-T~\cite{vmamba} &Cls.~\cite{vmamba} & 0.02 (0.06\%) &  61.23 (-26.25) &  54.36 (-43.46) &  61.62 (-28.60)\\
        VPT~\cite{prompt} & VMamba-T~\cite{vmamba}&Cls.~\cite{vmamba} & 0.03 (0.09\%)& 80.68 (-6.80)  & 89.23 (-8.59) & 80.34 (-9.78)\\
        \rowcolor{gray!20} Mamba-Adaptor & VMamba-T~\cite{vmamba}&Cls.~\cite{vmamba}  & 1.68 (5.56\%) & 86.82 (-0.66) & 93.24 (-4.58) & 86.45 (-3.76)  &\\
        \midrule[0.5pt]
         Full-Tuning &  VMamba-S~\cite{vmamba}&Cls.~\cite{vmamba}  & 55.25 (100\%) & 89.59 & 97.90 & 91.24\\
        Linear & VMamba-S~\cite{vmamba} &Cls.~\cite{vmamba} & 0.02 (0.04\%)&65.34 (-24.25) & 52.12 (-45.78) & 68.98 (-22.26) \\
       VPT~\cite{prompt} &   VMamba-S~\cite{vmamba}&Cls.~\cite{vmamba} &  0.03 (0.05\%) & 82.26 (-6.33)  &  82.48 (-15.42)&  64.23 (-27.01) & \\
      \rowcolor{gray!20} Mamba-Adaptor &  VMamba-S~\cite{vmamba} &Cls.~\cite{vmamba}  & 5.10 (9.25\%) & 88.30 (-1.39)  & 85.69 (-12.21)  & 82.33 (-8.91)\\
         \midrule[0.5pt]
       Full-Tuning &  VMamba-B~\cite{vmamba} &Cls.~\cite{vmamba}  & 95.36 (100\%) & 89.89 & 97.96 & 91.68\\
       Linear &   VMamba-B~\cite{vmamba} &Cls.~\cite{vmamba} & 0.03 (0.03\%)& 67.23 (-22.66) & 55.69 (-42.27)& 69.42 (-22.16) \\
      VPT~\cite{prompt} &  VMamba-B~\cite{vmamba} &Cls.~\cite{vmamba} &  0.04 (0.04\%)& 81.32 (-8.57) & 87.45 (-10.51) & 80.34 (-11.36) \\
        \rowcolor{gray!20}Mamba-Adaptor &  VMamba-B~\cite{vmamba} &Cls.~\cite{vmamba} & 6.8 (7.13\%)  & 88.34 (-1.5) & 88.52 (-8.44) & 83.21 (-8.47)\\
        \bottomrule[1pt]
    \end{tabular}}}
        \vspace{-2mm}
    \caption{Fine-tuning with the pre-trained base models for transfer learning tasks, where Mamba-Adaptor serves as an adaptor network. For tunable parameters, we report the percentage of the parameters. Additionally, we show both the absolute value and the gap value relative to the top 1 accuracy of the full-tuning setting. Cls. denotes normal image classificaiton training~\cite{swin, vmamba}.  }
    \vspace{-4mm}
    \label{tab:adaptor_cls}
\end{table*}

\subsection{Adaptor Network for Transfer Learning}
\label{sec:adaptor_exp}

\paragraph{Adaptor, Baseline, and Benchmark.}
We compare Mamba-Adaptor with three commonly used methods for fine-tuning.
(1) Linear probing: it involves adding an extra linear layer on top of the backbone and tuning the linear layer for evaluation.
(2) Full Fine-tuning: we unfreeze all model parameters and train them together.
(3) Visual Prompt Tuning (VPT)~\cite{prompt}: fine-tuning the additional visual tokens as prompt (we add 128 extra tokens).
We choose the VMamba~\cite{vmamba} with three variants, i.e., tiny/small/base, to evaluate the performance of the Adaptor. 
We also present the ViT-Base~\cite{vit} baseline network for comparison in Figure~\ref{tab:adaptor_cls}.
Following the setting in~\cite{adaptformer, prompt}, we choose three benchmarks for image recognition: CIFAR-100~\cite{cifar} includes 50,000 training images and 10K validation images at a resolution of 32×32 pixels across 100 labels; 
The Street View House Numbers (SVHN)~\cite{SVHN} dataset, used for digit classification, consists of over 600K labeled images, with 73K for training, 26K for testing, and 531K extra training images; The Food-101~\cite{food} dataset features 101 food categories, totaling 101K images.


\paragraph{Resutls on Transfer Learning.}
The results on three visual recognition benchmarks are reported in Table~\ref{tab:booster_cls}. 
Mamba-Adaptor maintains superior performance in top 1 classification accuracy across different datasets, compared to the other two finetuning methods, i.e., linear probe and VPT~\cite{prompt}.
Specifically, for the CIFAR-100 benchmark, Mamba-Adaptor-T/S/B preserves a 90\% performance with less than 5.56\%/9.25\%/7.13\% model parameters, compared to the full finetuning method, which corporates all the model parameters into the training. 
The performance of the linear probe and VPT~\cite{prompt} methods has a large gap between the full fine-tuning in all three visual recognition benchmarks~\cite{cifar, food, SVHN}, validating the effectiveness of our method.



\subsection{Ablation Studies}
\label{sec:abla_exp}

\begin{table}[t]
    \centering
    \setlength{\tabcolsep}{2.1mm}
    \subfloat{
    \scalebox{0.78}{
    \begin{tabular}{c | c c |c c c}
        \toprule[1pt]
          Setting & \makecell{VMamba-T } & \makecell{VMamba-S} & \makecell{Top1-acc\\(\%)} & \makecell{$AP^{b}$\\(\%)}& \makecell{$AP^{m}$\\(\%)}\\
        \midrule[0.5pt]
      \textcircled{1}& \XSolidBrush & \XSolidBrush & 77.2 &41.2&42.4 \\
             \textcircled{2} & static & \XSolidBrush &  77.6&41.6&42.9\\
       \textcircled{3} & \XSolidBrush & 1 scale &  77.5&41.4&42.7\\
       \textcircled{4} & static & 1 scale &  77.9&42.3&43.9\\
       \textcircled{5}  &learnbale  & 1 scale &  78.2& 42.9 &44.4\\
       \textcircled{6} & learnable & 2 scales & \textbf{78.4} &\textbf{43.2} &\textbf{44.8}\\
        \bottomrule[1pt]
    \end{tabular}}}
            \vspace{-2mm}
    \caption{Ablation study on the two functional modules of Mamba-Adaptor. The Mamba-Adaptor-b1 variant is adopted.  }
    \vspace{-2mm}
    \label{tab:abla1}
\end{table}

\paragraph{Adaptor-T/-S on Classification and Prediction tasks.}
Table~\ref{tab:abla2} shows the ablation studies on the two functional modules with different configurations, i.e., static/learnable state selection for Adaptor-T and multi-scale factor for the Adaptor-S. 
Compared to the base model without Mamba-Adaptor (\textcircled{1}), all the remaining settings (\textcircled{2} to \textcircled{6}) show improvement in varying degrees in both classification and dense prediction tasks. 
Notably, the learnable state selection scheme surpasses the static selection scheme with 0.3\% top-1 accuracy in ImageNet1k and 0.6\% in $AP^b$ in COCO object detection, validating the effectiveness of our dynamic selection scheme. 
We also observe a performance gain (0.7\% in $AP^m$) when adding additional depthwise kernels to the Adaptor-S due to the hierarchical representations.

\begin{table}[t]
    \centering
    \setlength{\tabcolsep}{1.mm}
    \subfloat{
    \scalebox{0.75}{
    \begin{tabular}{c | c c c| c c c c c}
        \toprule[1pt]
          Setting & Init. & Insert. & Weight S. &CIFAR-100 &SVHN &Food101 \\
        \midrule[0.5pt]
         \textcircled{1}& random & sequential  & \XSolidBrush & 62.34 &54.32 &65.12  \\
         \textcircled{2}& zero & parallel &  \XSolidBrush & 77.23 &67.98 & 78.32  \\
       \textcircled{3}& zero& sequential & \XSolidBrush  & 85.33 &92.43 & 83.43  \\
              \textcircled{4}& zero& parallel & \CheckmarkBold & \textbf{86.82} & \textbf{93.24} &\textbf{86.45} & \\
        \bottomrule[1pt]
    \end{tabular}}}
            \vspace{-2mm}
    \caption{Ablation study on different configurations for transfer learning. Weight S. denotes the weight-sharing coefficient in Adpator-T. Init. denotes the initialization method for the linear layer in Adaptor-T/-S. }
    \vspace{-2mm}
    \label{tab:abla2}
\end{table}

\paragraph{On Transfer Learning.}
As shown in Table~\ref{tab:abla2}, the initialization method matters for the transfer learning tasks. The zero initialization significantly raises the performance by 14.9/13.6/13.2 \% in top1 accuracy (\textcircled{2} vs. \textcircled{1}) in all three benchmarks compared to the random initialization. 
The parallel insertion form also contributes significantly to the performance in the transfer learning setting, as can be validated by the comparison between \textcircled{2} and \textcircled{3}. 
The potential reason is that zero initialization and the parallel insertion design maintain the original feature. 
It also matters for the booster module design as the pre-trained weight contains large-scale prior knowledge of the ImageNet~\cite{ImageNet}, and the extra Mamba-Adaptor module is of small model parameters.  
The weight-sharing coefficient strategy for Adaptor-T also improves the final performance while reducing the increased model parameters.

\begin{table}[t]
	\centering
	        \setlength{\tabcolsep}{2.1mm}	
	\subfloat{
    \scalebox{0.92}{
    \begin{tabular}{c | c c c c c}
        \toprule
                Transfer learning
              &   CIFAR 
              &   SVNH 
              &   Food101
        	 \\
        	\midrule
                ViM-Adaptor
                & +8.31
                & +6.27
        	& +9.18
        	\\
                VMamba-Bi-Adaptor
                & +6.42
                & +4.97
                & +7.02
        	\\\bottomrule
        \end{tabular}
	}}
	\vspace{-2mm}
	\caption{Transfer learning on Vision Mamba with bi-directional scanning, especially in comparison to the VPT~\cite{prompt} method.}
	\vspace{-2mm}
	\label{tab:abla_general}%
\end{table}

\paragraph{On Generalization ability.} We also conducted transfer learning experiments to show generalization ability. Tab.~\ref{tab:abla_general} shows that our method also shows good results on the Vision Mamba with bi-directional scanning~\cite{liu2024vmamba}.

\section{Conclusion and Limitations}
This paper presents a plug-and-play Mamba-Adaptor, which has three major usages. 
Mamba-Adaptor can serve as a general visual backbone for image classification and dense prediction tasks, which improves the baseline considerably. 
Moreover, it can also serve as a booster module that is inserted into a pre-trained vision backbone to enhance performance through extended training.
In transfer learning scenarios, the Mamba-Adaptor is an efficient module that enables the adaptation of a pre-trained Mamba backbone for various downstream visual recognition tasks.
Extensive experimental results evaluate the effectiveness of the Mamba-Adaptor. We hope that Mamba-Adaptor offers an insightful solution to enhance the Mamba scheme in the field of vision.
However, the scaling of the Mamba-Adaptor as a visual backbone remains open, and its applications to other variants of vision Mamba are part of future work.

\clearpage

{
    \small
    \bibliographystyle{ieeenat_fullname}
    \bibliography{ref}
}

\end{document}